\pdfoutput=1

\documentclass[letterpaper, 10 pt, conference]{ieeeconf}  

\usepackage[utf8]{inputenc}
\usepackage{graphicx}
\usepackage{hyperref}
\usepackage{subcaption}
\usepackage{amsmath,amssymb} 
\usepackage{color}

\newif\ifdraft
\draftfalse

\definecolor{orange}{rgb}{1,0.5,0}
\definecolor{violet}{RGB}{70,0,170}
\definecolor{magenta}{RGB}{170,0,170}
\definecolor{dgreen}{RGB}{0,150,0}

\usepackage[normalem]{ulem}

\newcommand{\comment}[1]{}

\newcommand{\bH}{\mathbf{H}}

\newcommand{\OURS}[0]{\textbf{OURS}}

\newcommand{\colvec}[3]{\ensuremath{
		\begin{bmatrix}{#1}	\\	{#2}	\\	{#3} \end{bmatrix}
}}

\newcommand{\tr}{^\intercal}

\definecolor{tab_highlight}{rgb}{0.95,0.95,0.95}

\IEEEoverridecommandlockouts                              

\title{\LARGE \bf
Geometric and Physical Constraints for \\ Drone-Based Head Plane Crowd Density Estimation
}

\author{Weizhe Liu, Krzysztof Lis, Mathieu Salzmann, Pascal Fua
\thanks{The authors are  with the Computer Vision Laboratory, School of Communication and Computer Sciences, École Polytechnique Fédérale de Lausanne (EPFL), Switzerland.}
\thanks{This work was supported in part by a grant from the Swiss Federal Office for Defense Procurement.}
}

\begin{document}

\maketitle
\thispagestyle{empty}
\pagestyle{empty}

\begin{abstract}

State-of-the-art methods for counting people in crowded scenes rely on deep networks to estimate crowd density in the image plane. 
While useful for this purpose, this image-plane density has no immediate physical meaning because it is subject to perspective distortion. This is a concern in sequences acquired by drones because the viewpoint changes often.
This distortion is usually handled implicitly by either learning scale-invariant features or estimating density in patches of different sizes, neither of which accounts for the fact that scale changes must be consistent over the whole scene. 

In this paper, we explicitly model the scale changes and reason in terms of people per square-meter. We show that feeding the perspective model to the network allows us to enforce global scale consistency and that this model can be obtained on the fly from the drone sensors.
In addition, it also enables us to enforce physically-inspired temporal consistency constraints that do not have to be learned.
This yields an algorithm that outperforms state-of-the-art methods in inferring crowd density from a moving drone camera especially when perspective effects are strong.

\end{abstract}


\section{Introduction}

With the growing prevalence of drones, drone-based crowd density estimation becomes increasingly relevant to applications such as autonomous landing and video surveillance. In recent years, the emphasis has been on developing {\it counting-by-density} algorithms that rely on regressors trained to estimate the density of crowd per unit area so that the total numbers of people can be obtained by integration, without explicit detection 
being required. The regressors can be based on Random Forests~\cite{Lempitsky10}, Gaussian Processes~\cite{Chan09}, or more recently  Deep Nets~\cite{Zhang15c,Zhang16s,Onoro16,Sam17,Xiong17,Sindagi17,Shen18,Liu18a,Li18e,Sam18,Shi18,Liu18b,Idrees18,Ranjan18,Cao18}, 
with most state-of-the-art approaches now relying on the latter.


\begin{figure}
\centering
\begin{tabular}{ccc}
	\includegraphics[width=.3\linewidth]{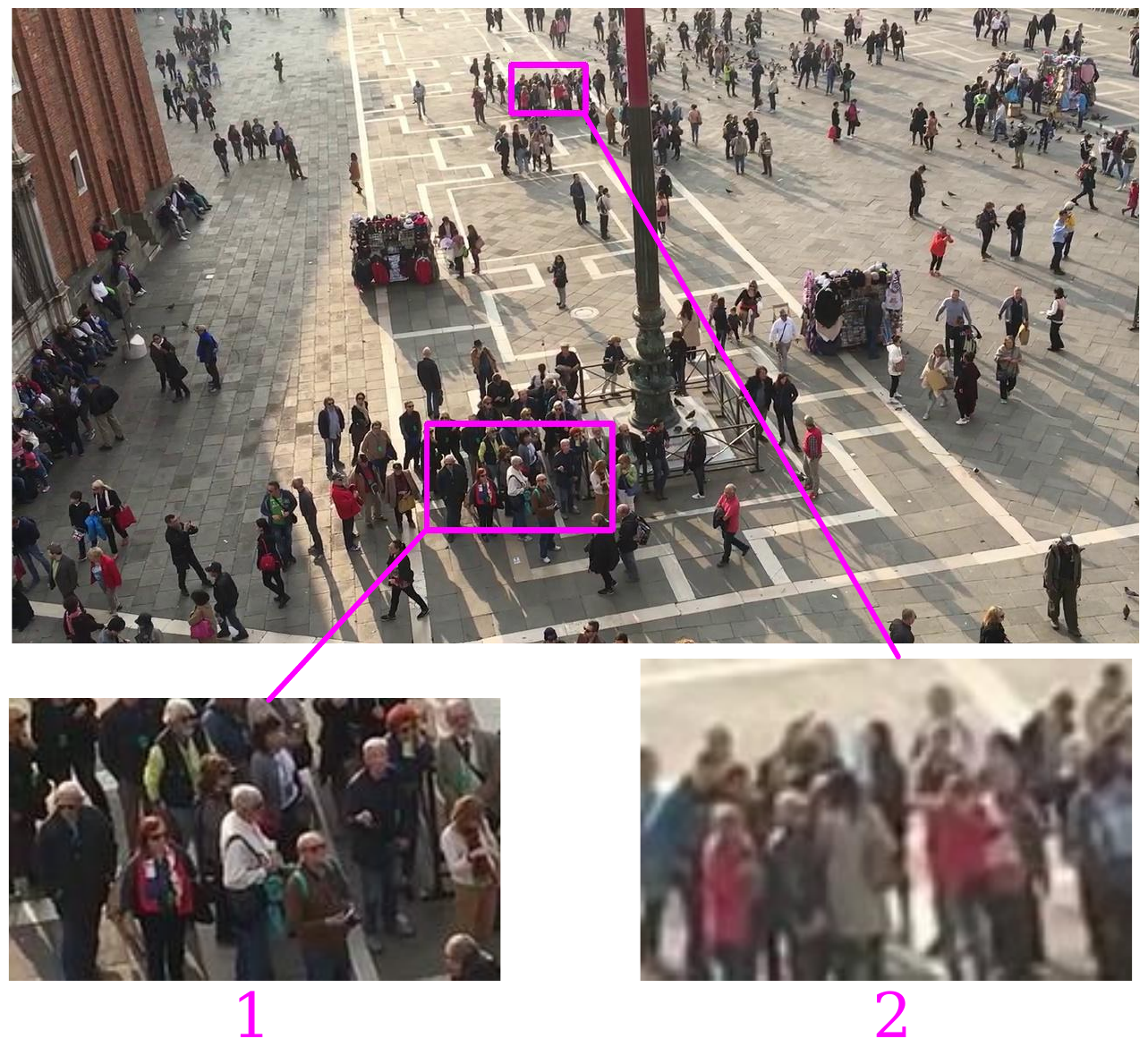}&
	\includegraphics[width=.3\linewidth]{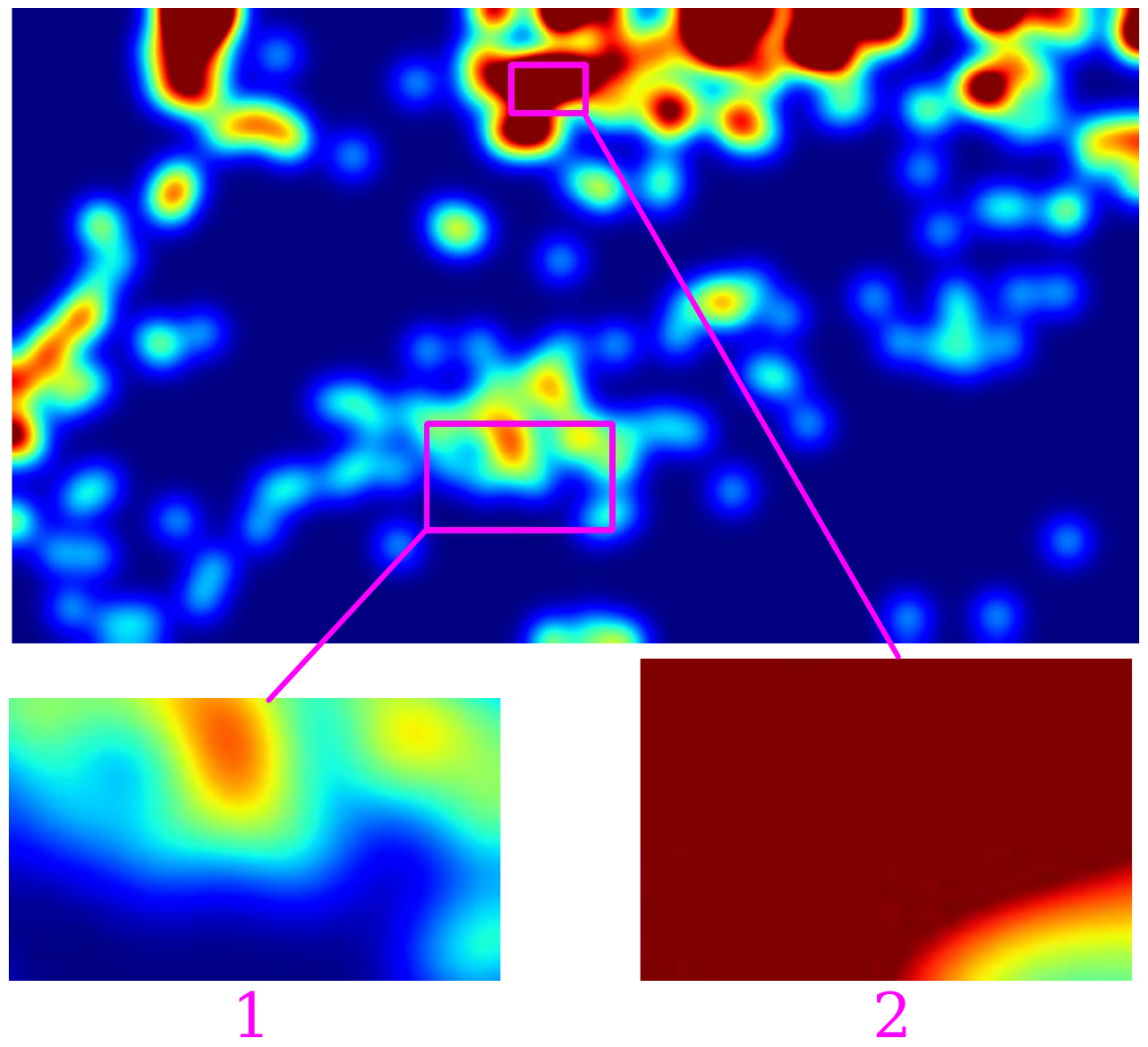}& 
	\includegraphics[width=.3\linewidth]{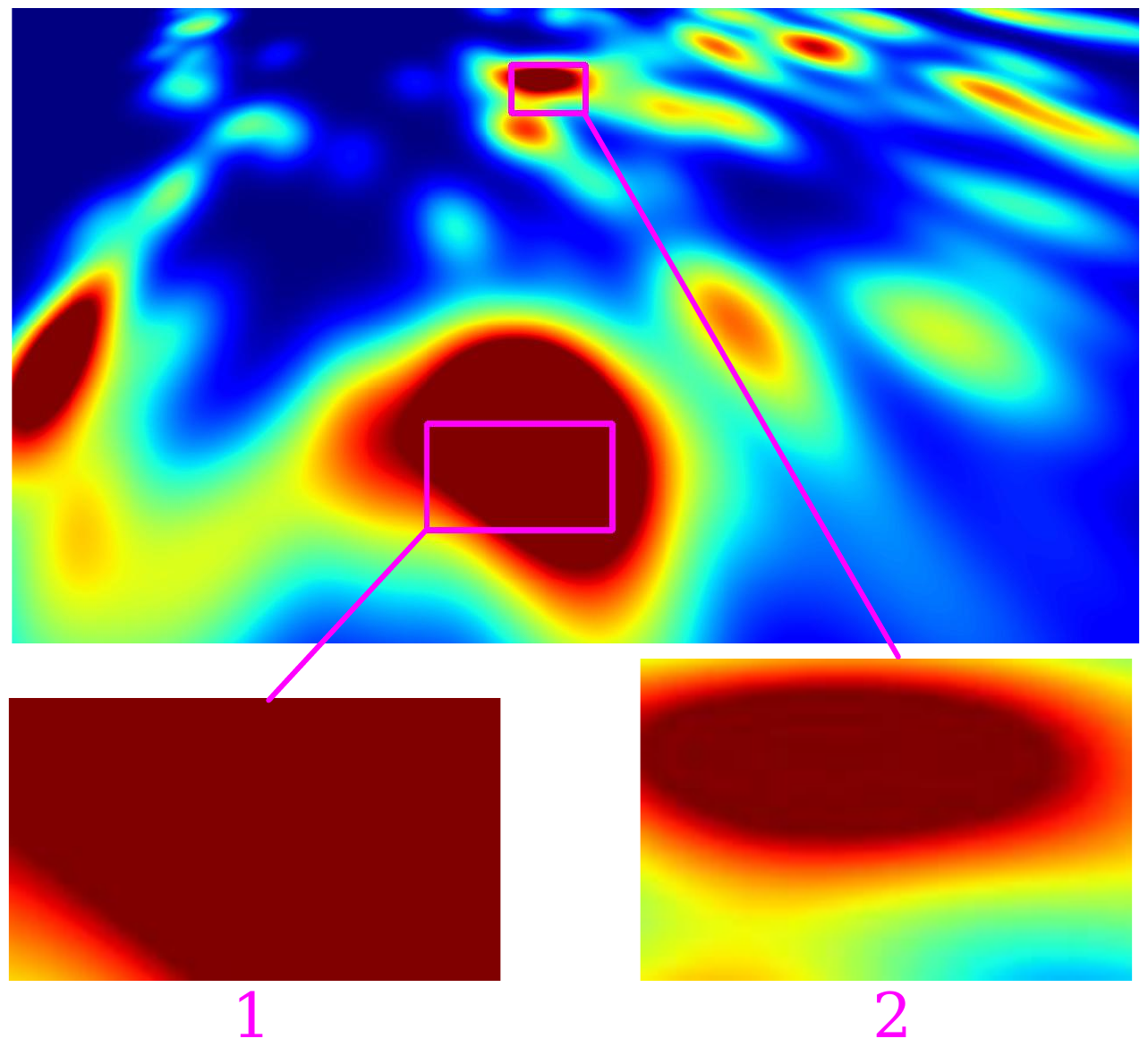} \\
	(a)&(b)&(c)
\end{tabular}
\vspace{-3mm}
\caption{
	{\bf Measuring people density.} {\small 
	{\bf (a)} An image of Piazza San Marco in Venice. The two purple boxes highlight patches in which the crowd density per square meter is similar. 
	{\bf (b)} Ground-truth {\it image density} obtained by averaging the head annotations in the image plane. The two patches are in the same locations as in (a). The density per square pixel strongly differs due to perspective distortion: the farther patch 2 wrongly features a higher density than closer patch 1, even though the people do not stand any closer to each other.
	{\bf (c)} By contrast the ground-truth {\it head plane density} introduced in Section~\ref{sec:headDensity} is unaffected by perspective distortion. The density in the two patches now has similar peak values, as it should.
}}

\label{fig:density}
\end{figure}

While effective, these algorithms all estimate density in the image plane. As a consequence, and as can be seen in Fig.~\ref{fig:density}(a,b), two regions of the scene containing the same number of people per square meter can be assigned different densities.
However, for the purposes of autonomous landing or crowd size estimation, the density of people on the ground is a more relevant measure and  is {\it not} subject to such distortions, as shown in Fig.~\ref{fig:density}(c).

In this paper, we therefore introduce a crowd density estimation method that \emph{explicitly} accounts for perspective distortion to produce a real-world density map, as opposed to an image-based one. To this end, it takes advantage of the fact that drone cameras can be naturally registered to the scene using the drone's internal sensors, which as we will see are accurate enough for our purposes. This contrasts with methods that \emph{implicitly} deal with perspective effects by either learning scale-invariant features~\cite{Zhang16s,Sam17,Sindagi17} or estimating density in patches of different sizes~\cite{Onoro16}. Unlike these, we model perspective distortion globally and account for the fact that people's projected size changes consistently across the image. To this end, we feed to our density-estimation CNN not  only the original image but also an identically-sized image that contains the local scale, which is a function of the camera orientation with respect to the ground plane.

An additional benefit of reasoning in the real world is that we can encode physical constraints to model the motion of people in a video sequence. Specifically, given a short sequence as input to our network, we impose temporal consistency by forcing the densities in the various images to correspond to physically possible people flows. In other words, we \emph{explicitly} model the motion of people, with physically-justified constraints, instead of \emph{implicitly} learning long-term dependencies only across annotated frames, which are typically sparse over time, via LSTMs, as is commonly done in the literature~\cite{Xiong17}.

Our contribution is therefore an approach that incorporates geometric and physical constraints directly into an end-to-end learning formalism for crowd counting using information directly obtained from the drone sensors. As evidenced by our experiments, this enables us to outperform the state-of-the-art on a drone-based video sequences with severe perspective distortion.


\section{Related work}

Early crowd counting methods~\cite{Wang11a,Lin10} tended to rely on {\it counting-by-detection}, that is, explicitly detecting individual heads or bodies  and then counting them. Unfortunately, in very crowded scenes, occlusions make detection 
difficult, and these approaches have been largely displaced by {\it counting-by-density-estimation} ones, which rely on training a regressor to estimate people density in various parts of the image and then integrating. This trend 
started with~\cite{Lempitsky10}, and~\cite{Chan09}, which used Random Forests and Gaussian Process regressors, respectively. 

Even though approaches such as these early ones that rely on low-level features---a survey of which can be found  in~\cite{Sindagi17}---can be effective they have now mostly been superseded by CNN-based methods. The same can be said about methods that count objects instead of people~\cite{Arteta14,Arteta16}.

\paragraph{\bf Perspective Distortion}

Earlier approaches to handling such distortions~\cite{Zhang15c} involve regressing to both a crowd count and a density map. Unlike ours that passes a perspective map as an input to the deep network, they use the perspective map to compute a metric and use it to retrieve candidate training scenes with similar distortions before tuning the model. This complicates training, which is not end-to-end, and decreases performance.

These approaches were recently extended by~\cite{Sam17}, whose {\it SwitchCNN} exploits a classifier that greedily chooses the 
sub-network that yields the best crowd counting performance. Max pooling is used extensively to down-scale the density map output, 
which improves the overall accuracy of the counts but decreases that of the density maps as pooling incurs a loss in localization precision.

Perspective distortion is also addressed in~\cite{Onoro16} via a scale-aware model called {\it HydraCNN}, 
which uses different-sized patches as input to the CNN to achieve scale-invariance. To the same end,
different kernel sizes are used  in~\cite{Zhang16s} and  in~\cite{Idrees18} features from different layers are extracted instead.
In the recent method of~\cite{Sindagi17}, a network dubbed CP-CNN combines local and global 
information obtained by learning density at different resolutions. It also accounts for density map quality by adding extra information 
about the pre-defined density level of different patches and images.
While useful, this information is highly scene specific and would make generalization difficult. 
More recent works use different techniques, such as a growing CNN~\cite{Sam18}, fusing crowd counting with people detection~\cite{Liu18a}, adding a new measurement between prediction and ground truth density map~\cite{Cao18}, using a scale-consistency regularizer~\cite{Shen18}, employing a pool of decorrelated regressors~\cite{Shi18}, refining the density map in an iterative process~\cite{Ranjan18}, leveraging web-based unlabeled data~\cite{Liu18b}, to further boost performance. However, none of them is specifically designed to handle perspective effects.

In any event, all the approaches mentioned above rely on the network learning about perspective effects without explicitly modeling them. 
As evidenced by our results, this is suboptimal given the finite amounts of data available in practical situations. Furthermore, while learning about perspective effects to account for the varying people sizes, these methods still predict density in the image plane, thus leading to the unnatural phenomenon that real-world regions with the same number of people are assigned different densities, as shown in Fig.~\ref{fig:density}(b). By contrast, we produce densities expressed in terms of number of people per square meter of ground, such as the ones shown in Fig.~\ref{fig:density}(c), and thus are immune to this problem.

\paragraph{\bf Temporal Consistency}

The recent method of~\cite{Xiong17} is representative of current approaches in enforcing temporal consistency by incorporating an 
LSTM module~\cite{Hochreiter97}  to perform feature fusion over time. This helps but can only capture temporal 
correlation across annotated frames, which are widely separated in most existing training sets. In other words, 
the network can only learn long-term dependencies at the expense of shorter-term ones. 

By contrast, since we reason about crowd density in real world space, we can model physically-correct temporal dependencies via frame-to-frame feasibility constraints, without any additional annotations.


\section{Perspective Distortion}
\label{sec:perspective}

All existing approaches estimate the crowd density in the image plane and in terms of people per square pixel,  which changes across the image even if the true crowd density per square meter is constant.
For example, in many scenes such as the one of  Fig.~\ref{fig:density}(a), the people density in farther regions is higher than that in closer regions, as can be seen in  Fig.~\ref{fig:density}(b). 

In this work, we train the system to directly predict the crowd density in the physical world, which does not suffer from this problem and is therefore unaffected by perspective distortion, assuming that people are standing on an approximately flat surface. Our approach could easily be extended to a non flat one given a terrain model.
In a crowded scene, people's heads are more often visible than their feet. Consequently, it is a common practice to provide annotations in the form of a dot on the head for supervision purposes. To account for this, we define a {\it head plane}, parallel to the ground and lifted above it by the average person's height.
We assume that the camera has been calibrated so that we are given the homography between the image and the head plane. 

\subsection{Image Plane versus Head Plane Density}
\label{sec:headDensity}

Let $\bH_i$ be the homography from an image $I_i$ to its corresponding head plane. We define the ground-truth density as a sum of Gaussian kernels
centered on peoples' heads in the head plane. Because we work in the physical world, we can use the same kernel size across the entire scene and across all scenes.
A head annotation $P_i$, that is, a 2D image point expressed in projective coordinates, is mapped to $\bH_i P_i$ in the head plane.
Given a set $A_{i}=\{P_i^1,...,P_i^{c_i}\}$ of $c_i$ such annotations, we take the {\it head plane density} 
$G_i'$ at point $P'$ expressed in head plane coordinates to be
\begin{equation}
	G_i'(P')= \sum_{j=1}^{c_i} \mathcal{N}(P'; \bH_i P_i^j, \sigma) \; ,
\end{equation}
where $\mathcal{N}(. ; \mu,\sigma)$ is a 2D Gaussian kernel with mean $\mu$ and variance $\sigma$. We can then map this head plane density to the image coordinates, which yields a density at pixel location $P$ given by
\begin{equation}
	G_i(P) = G_i'(\mathbf{H}_i  P)  \;  .
\end{equation}
An example density $G_i$ is shown in Fig.~\ref{fig:density}(c). Note that, while the density is Gaussian in the head plane, it is {\it not} in the image plane.

\subsection{Geometry-Aware Crowd Counting}
\label{sec:crowdCounting}

Since the head plane density map can be transformed into an image of the same size as that of the original image, we could simply train a deep network to take a 3-channel RGB image as input and output the corresponding density map. However, this would mean neglecting the geometry encoded by the ground plane homography, namely the fact that the local scale does not vary arbitrarily across the image and must remain globally consistent. 

\begin{figure}
  \centering
  \includegraphics[width=\linewidth]{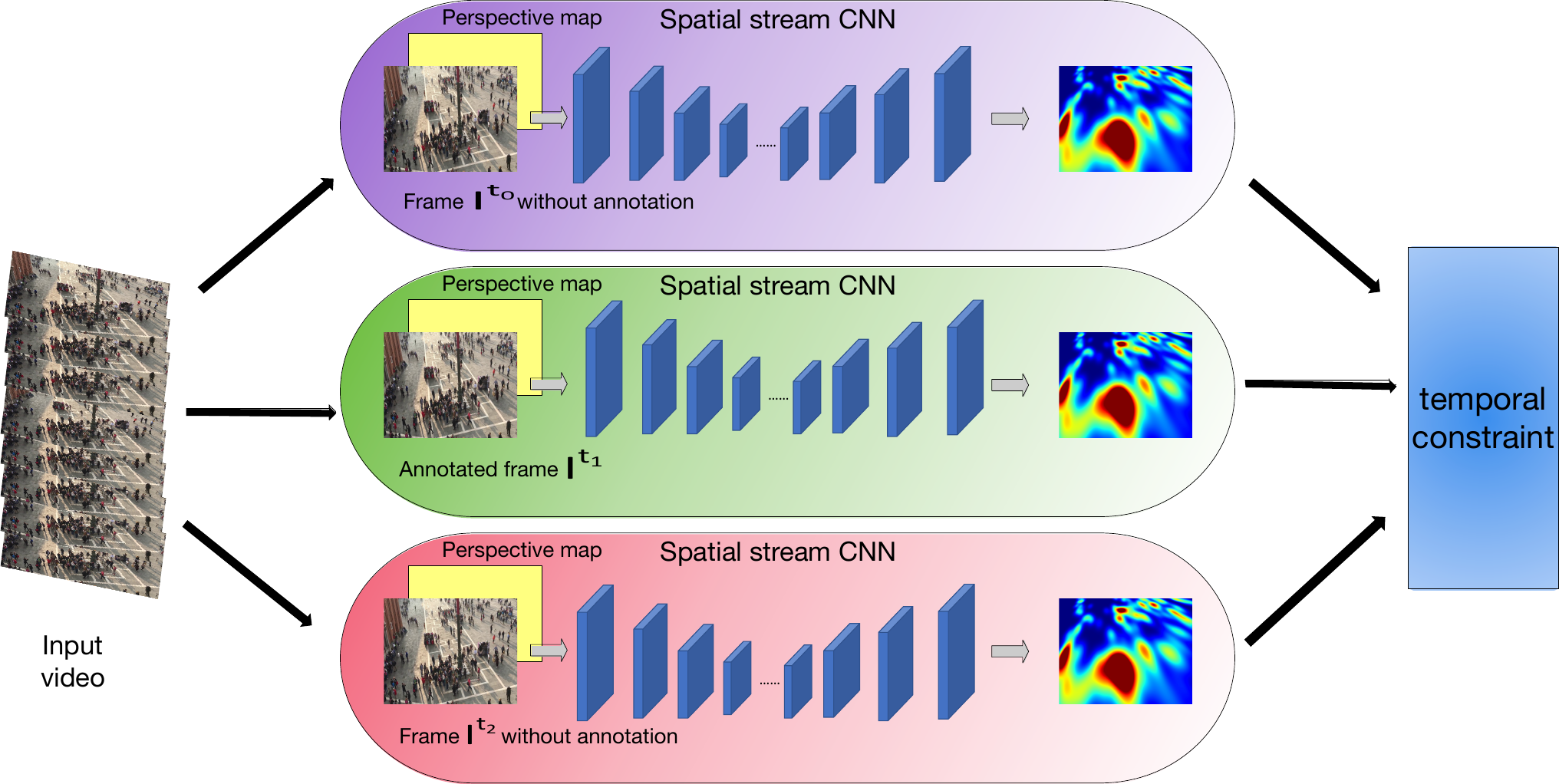}
\caption{{\bf Three-stream architecture.} {\small A spatial stream is a CSRNet~\cite{Li18e} with 3 transposed convolutional layers, that takes as input  the image and a perspective map. It is duplicated three times to process images taken at different times and minimize a loss that enforces temporal consistency constraints.}}
\label{fig:architecture}
\end{figure}

To account for this, we associate to each image $I$ a {\it perspective map} $M$ of the same size as $I$ containing the local scale of each pixel, that is, the factor by which a small area around the pixel is multiplied when projected to the head plane. We then use a convolutional network with 4 input channels instead of only 3. The first three are the usual RGB channels, while the fourth contains the perspective map. We will show in the result section that this substantially increases accuracy over using the RGB channels only.
This network is one of the {\it spatial streams} depicted by Fig.~\ref{fig:architecture}.
To learn its weights $\Theta$, we minimize the {\it head plane loss} $L_{H}(I,M,G;\Theta)$,
which we take to be the mean square error between the predicted head plane density and the ground-truth one.


To compute the perspective map $M$, let us first consider the image pixel $(x, y)\tr$ and an infinitesimal area $dx\;dy$ surrounding it. Let $(x',y')\tr$ and $dx' dy'$ be their respective projections on the head plane. We take $M(x,y)$, the scale at $(x,y)\tr$, to be $(dx'dy') / (dx\;dy)$, which we compute as follows. 
Using the variable substitution equation, we write
\begin{equation}
	dx' dy' = |\det(J(x, y))| dx \; dy \; ,
\end{equation}
where $J(x, y)$ is the Jacobian matrix of the coordinate transformation at the point $(x, y)\tr$:
\begin{equation}
J = 
\begin{bmatrix}
	\frac{\partial x'}{\partial x} & \frac{\partial x'}{\partial y} \\
	\frac{\partial y'}{\partial x} & \frac{\partial y'}{\partial y} \\	
\end{bmatrix}
.
\end{equation}
The scale map $M$ is therefore equal to
\begin{equation}
M(x, y) =  |\det(J(x, y))| .
\label{eq:localScale}
\end{equation}
The detailed solution can be found in~\cite{Chum10}.
Eq.~\ref{eq:localScale} enables us to compute the perspective map that we use as an input to our network, as discussed above. It also allows us to convert between people density $F$ in image space, that is, people per square pixel, and people density $G'$ on the head plane. More precisely, let us consider a surface element $dS$ in the image around point $(x, y)\tr$. It is scaled by $\bH$ into $dS' = M(x, y) dS$. Since the projection does not change the number of people, we have 
\begin{eqnarray}
	F(x, y) dS & = & G'(x', y')  dS' = G'(x', y') M(x, y) dS \nonumber \\
	\Rightarrow F(x, y) & = & M(x, y) G'(x', y') \; .
\end{eqnarray}
Expressed in image coordinates, this becomes
\begin{equation} \label{eq:1}
	F(x, y) = M(x, y) G(x, y) \; ,
\end{equation}
which we use in the results section to compare our algorithm that produces head plane densities against the baselines that estimate image plane densities. 

\subsection{Obtaining scene geometry from UAV sensors}

We calculate the homography matrix $\mathbf{H}$ using the camera's altitude $h$ and pitch angle $\theta$ reported by the UAV sensors. We choose the world coordinate frame such that the head plane is given by $Z=0$ and the origin $(0, 0, 0)\tr$ is directly under the UAV. The camera extrinsics are described by the rotation matrix $R = R_y(\frac{\pi}{2} + \theta)$ and translation vector $t = (0, 0, h)\tr$.

The relation between a point $(x_h, y_h, 0)\tr$ on the head plane and its projection $(u, v)\tr$ onto the image is expressed by the following equation, in homogenous coordinates:
\begin{equation}\label{eq:plane_to_cam}
\colvec{u}{v}{1}
=
K
\begin{bmatrix}
R_{11}	&	R_{12}	&	R_{13}	&	t_1	\\
R_{21}	&	R_{22}	&	R_{23}	&	t_2	\\
R_{31}	&	R_{32}	&	R_{33}	&	t_3	\\
\end{bmatrix}
\left[
\begin{array}{c}
x_h	\\	y_h	\\ 0	\\ 1
\end{array}
\right]
, 
\end{equation}
where $K$ is the camera's intrinsic matrix and $w \neq 0$ is an arbitrary scale factor.
Solving for $(x_h, y_h)\tr$ we obtain:
\begin{equation}\label{eq:cam_to_plane}
\colvec{x_p}{y_p}{1}
=
w 
\left( 
K
\begin{bmatrix}
R_{11}	&	R_{12}	&	t_1	\\
R_{21}	&	R_{22}	&	t_2	\\
R_{31}	&	R_{32}	&	t_3	\\
\end{bmatrix}
\right)
^{-1}
\colvec{u}{v}{1}
.
\end{equation}
The transformation from the image to the head plane is therefore given by the homography 
$\mathbf{H}=\left( 
K
\left[ \begin{array}{c|c|c}
R_{1}	&	R_{2}	&	t	\\
\end{array} \right]
\right) 
^{-1}$.

\section{Temporal Consistency}
\label{sec:temporal}

The spatial stream network depicted at the top of Fig.~\ref{fig:architecture} operates on single frames of a video sequence. To increase robustness, we now show how to enforce temporal consistency across triplets of frames. Unlike in an LSTM-based approach, such as~\cite{Xiong17}, we can do this across any three frames instead of only across annotated frames. Furthermore, by working in the real world plane instead of the image plane, we can explicitly exploit physical constraints on people's motion.

\subsection{People Conservation}
\label{sec:constraints}

An important constraint is that people do not appear or disappear from the head plane except at the edges or at specific points that can be marked as exits or entrances. To model this, we partition the head plane into $K$ blocks. Let $N(k)$ for ${1\leq k \leq K}$ denote the neighborhood of block $B_k$, including $B_k$ itself. Let $m_k^t$ be the number of people in $B_k$ at time $t$ and let $t_0 < t_1 < t_2$ be three different time instants. In experiments, we empirically set the block size to 30 by 30 pixels.

If we take the blocks to be large enough for people not be able to traverse more than one block between two time instants, people in the interior blocks can only come from a block in $N(k)$ at the previous instant and move to a block in $N(k)$ at the next. As a consequence, we can write
\begin{equation}
    \forall k \quad  m_{k}^{t_1}  \le \sum_{i \in N(k)} m_{i}^{t_0} \mbox{ and } m_{k}^{t_1}  \le \sum_{i \in N(k)}m_{i}^{t_2} \; . \label{eq:conservation}
\end{equation}
In fact, an even stronger equality constraint could be imposed  as in~\cite{Berclaz11} by explicitly modeling people flows from one block to the next with additional variables predicted by the network. However, not only would this increase the number of variables to be estimated, but it would also require enforcing hard constraints between different network's outputs.

In practice, since our networks output head plane densities, we write
\begin{equation}
	m_{k}^{t} = \sum_{(x',y')\tr\in B_k} \hat{G}^{'t}(x',y') \; ,
\label{eq:sumDensity}
\end{equation}
where $\hat{G}^{'t}$ is the predicted people density at time $t$, as defined in Section~\ref{sec:crowdCounting}. This allows us to reformulate the constraints of Eq.~\ref{eq:conservation} in terms of densities. 

\subsection{Siamese architecture}
\label{sec:siamese}

To enforce these constraints, we introduce the siamese architecture depicted by Fig.~\ref{fig:architecture}, with weights $\Theta$. It comprises three identical streams, each stream is a CSRNet~\cite{Li18e} with 3 transposed convolutional layers added before the last convolutional layer, so that the input image and output density map have the same size.  These three identical steams take as input images acquired at times $t_0$, $t_1$, and $t_2$  along with their corresponding perspective maps, as described in Section~\ref{sec:crowdCounting}. Each one produces a head plane density estimate $G^{'t_i}$ and we define the temporal loss term $L_{T}(I^{t_0}\hspace{-1mm}, I^{t_1}\hspace{-1mm},I^{t_2}\hspace{-1mm},M^{t_0}\hspace{-1mm},M^{t_1}\hspace{-1mm},M^{t_2}; \Theta)$ as
\begin{small}
\begin{equation}
     \frac{1}{2K}  \sum_{k=1}^{K} [ (max(0, m_{k}^{t_1} - U_{k}^{t_0}))^{2} +  (max(0, m_{k}^{t_1} - U_{k}^{t_2}))^{2}] \; ,
\end{equation}
\end{small}
where $m_{k}^t$ is the sum of predicted densities in block $B_k$, as in Eq.~\ref{eq:sumDensity}, and $U_{k}^{t}= \sum_{i \in N(k)}  m_{i}^{t} $  is the sum of densities in the neighborhood of $B_k$.

In other words, $L_T$ penalizes violations of the constraints of Eq.~\ref{eq:conservation}. 
At training time, we minimize the composite loss 

\begin{small}
\begin{equation}
 L_{H}(I^{t_1}\hspace{-1mm} ,M^{t_1}\hspace{-1mm},G'^{t_1};\Theta) \hspace{-1mm} + \hspace{-1mm} L_{T}(I^{t_0}\hspace{-1mm} , I^{t_1}\hspace{-1mm},I^{t_2}\hspace{-1mm},M^{t_0}\hspace{-1mm},M^{t_1}\hspace{-1mm},M^{t_2}; \Theta)  ,
\end{equation}
\end{small}
where $L_{H}$ is the head plane loss introduced in Section~\ref{sec:crowdCounting}. 
Since the loss requires the ground truth density only for frame $I^{t_1}$, we only need annotations for that frame. Therefore, we can use arbitrarily-spaced and unannotated frames to impose temporal consistency and improve robustness, which is not something LSTM-based methods can do.  


\newcommand{\csr}[0] {{\bf CSRNet}}
\newcommand{\mcnn}[0] {{\bf MCNN}}
\newcommand{\switch}[0] {{\bf SwitchCNN}}

\newcommand{\nogeom}[0]{{\bf OURS-NoGeom}}
\newcommand{\notemp}[0]{{\bf OURS-GeomOnly}}
\newcommand{\ours}[0]{{\bf OURS}}

\newcommand{\venice}[0] {{\bf Venice}}
\newcommand{\shangai}[0] {{\bf ShangaiTech}}
\newcommand{\expo}[0] {{\bf WorldExpo}}

\section{Experiments}

\label{sec:experiments}

\subsection{Datasets and Experimental Setup}

Our approach is designed to handle perspective effects as well as to enforce temporal consistency. As there is no publicly available drone-based crowd counting dataset, we filmed a six-minute long sequence using a DJI phantom 4 pro drone flying over a university campus and filming it from many different perspectives. We manually annotated 90 images such as the one of Fig.~\ref{fig:exp_drone} and used 54 of them for training and validation purposes and the remainder for testing. The people count ranges from 54 to 301 in this dataset. We will refer to it as \textbf{Campus}. In the supplementary material, we provide a video showing our results on a subsequence.

To demonstrate that our approach also works in a very different context, we also evaluate it on the publicly available \textbf{Venice}~\cite{Liu19a} dataset, which was recorded using a mobile phone. It features Piazza San Marco as seen from various viewpoints on the second floor of the basilica and substantial perspective effects. This dataset comprises 4 different sequences and 167 annotated frames. Fig.~\ref{fig:density} depicts one of these. The white lines on the Piazza make it easy to estimate the plane homography using standard photogrammetric techniques and the sequence is thus a good proxy for drone-acquired footage.

We focus on head-plane and ground-plane densities, as opposed to image-plane densities, because they are the ones that have a true physical meaning independently of the camera motion. In this section, we therefore report our results and baselines ones in head-plane density terms. However, we also provides image plane density results to demonstrate that our model outperforms the baselines in both cases.

\subsection{Baselines}

We benchmark our approach against three recent methods for which the code is publicly available: \csr{}~\cite{Li18e}, \mcnn{}~\cite{Zhang16s} and \switch{}~\cite{Sam17}.  As discussed in the related work section, they are representative of current approaches to handling the fact that people's sizes vary depending on their distance to the camera. 

We will refer to our complete approach as \ours{}. To tease out the individual contributions of its components, we also evaluate two degraded versions of it. \nogeom{} uses the CNN to predict densities but does not feed it the perspective map as input. \notemp{}  uses the full approach described in Section~\ref{sec:perspective} but does not impose temporal consistency.

\subsection{Evaluation Metrics}
\label{sec:metrics}

Most previous works in crowd density estimation use mean absolute error (MAE) and root mean squared error (RMSE) as their evaluation metric. They are defined as  
\begin{small}
\begin{equation}
 \hspace{0mm} \mbox{MAE} = \hspace{-1mm}\frac{1}{N}\sum_{1}^{N}|z_{i}-\hat{z_{i}}| \mbox{ and } \mbox{RMSE} =\hspace{-1mm} \sqrt{\frac{1}{N}\sum_{1}^{N}(z_{i}-\hat{z_{i}})^{2}} ,
\end{equation}
\end{small}
where $N$ is the number of test images, $z_{i}$ denotes the true number of people inside the ROI of the $i$th image and $\hat{z_{i}}$ the estimated number of people. 
While indicative, these two metrics are very coarse, since these two metrics only take into consideration the total number of people irrespective of where in the scene they may be,
so they are incapable of evaluating the correctness of the spatial distribution of crowd density.
A false positive in one region, coupled with a  false negative in another, can still yield a perfect total number of people.

We therefore introduce one additional metric that provide finer grained measures, accounting for localization errors. We name it the mean pixel-level absolute error (MPAE) and take it to be
\begin{small}
\begin{equation}
   \mbox{MPAE} = \hspace{-1mm} \frac{\sum_{i=1}^{N}\hspace{-1mm}\sum_{j=1}^{H}\hspace{-1mm}\sum_{k=1}^{W}|D_{i,j,k}\hspace{-1mm}-\hspace{-1mm}\hat{D}_{i,j,k}| \hspace{-1mm}\times \hspace{-1mm}{\bf 1}_{ \{D_{i,j,k }\in R_{i}\} } }{N} , 
\end{equation}
\end{small}
where $D_{i,j,k}$ is the ground-truth density of the $i$th image at pixel $(j,k)$, $\hat{D}_{i,j,k}$ is the corresponding estimated density, $R_{i}$ is the ROI of the $i$th image, ${\bf 1}_{ \{\cdot\} }$ is the indicator function, and $W$ and $H$ are the image dimensions. MPAE quantifies how wrongly localized the densities are. 

The baseline models~\cite{Zhang16s,Li18e,Sam17} are designed to predict density in the image plane instead of the head plane, as our model does. Fortunately, the densities in image plane and head plane can be easily converted into each other, as shown in Section~\ref{sec:perspective}. For a fair comparison, we therefore train the baseline models~\cite{Zhang16s,Li18e,Sam17} as reported in original the papers to estimate density in the image-plane. We then used Eq.~\ref{eq:1}  to convert to head-plane density. Thus we can use the MAE, RMSE, and MPAE metrics to compare both kinds of densities. 

\subsection{Quantitative Evaluation}

\begin{figure*}
\begin{center}
  \begin{tabular}{cccc}
    \includegraphics[width=.22\linewidth]{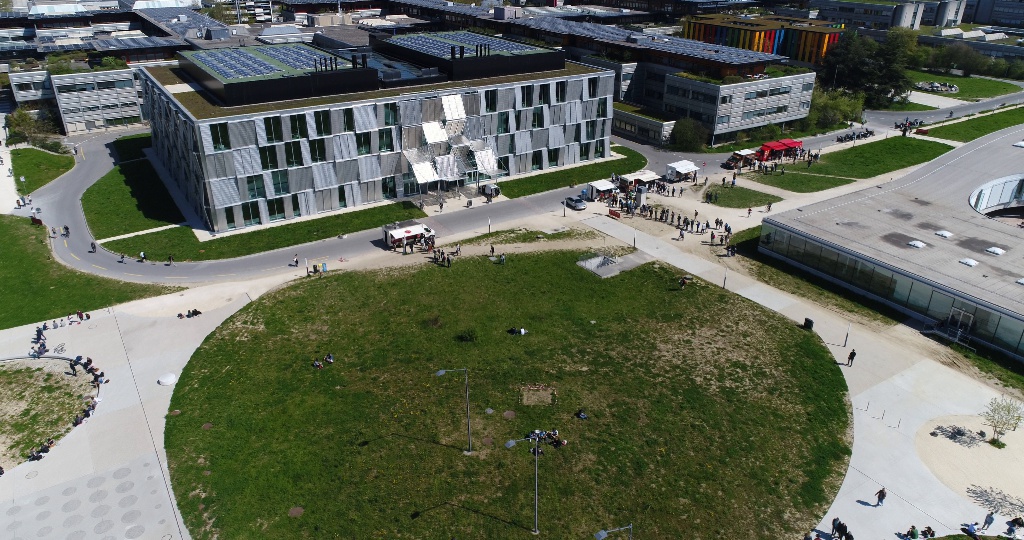}&
    \includegraphics[width=.22\linewidth]{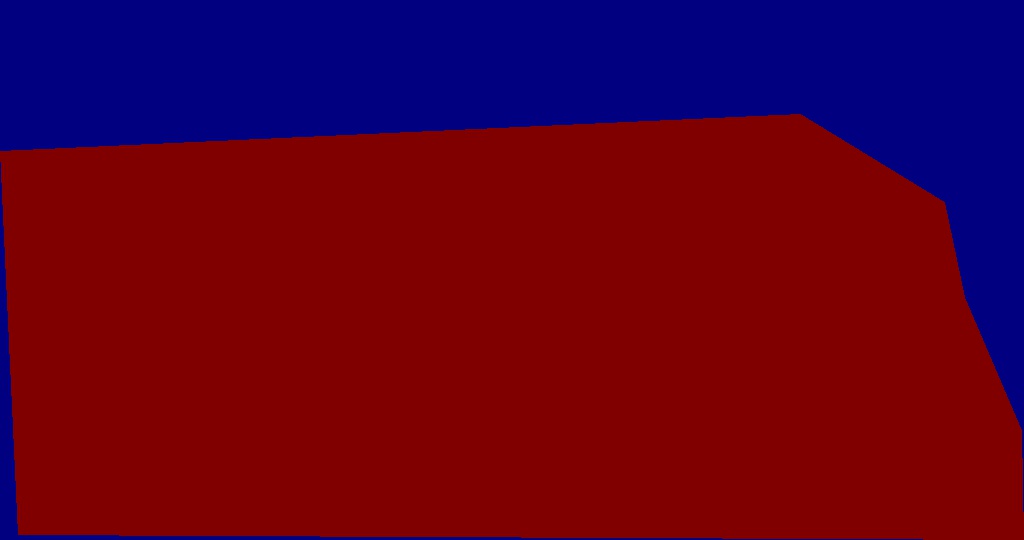}&
    \includegraphics[width=.22\linewidth]{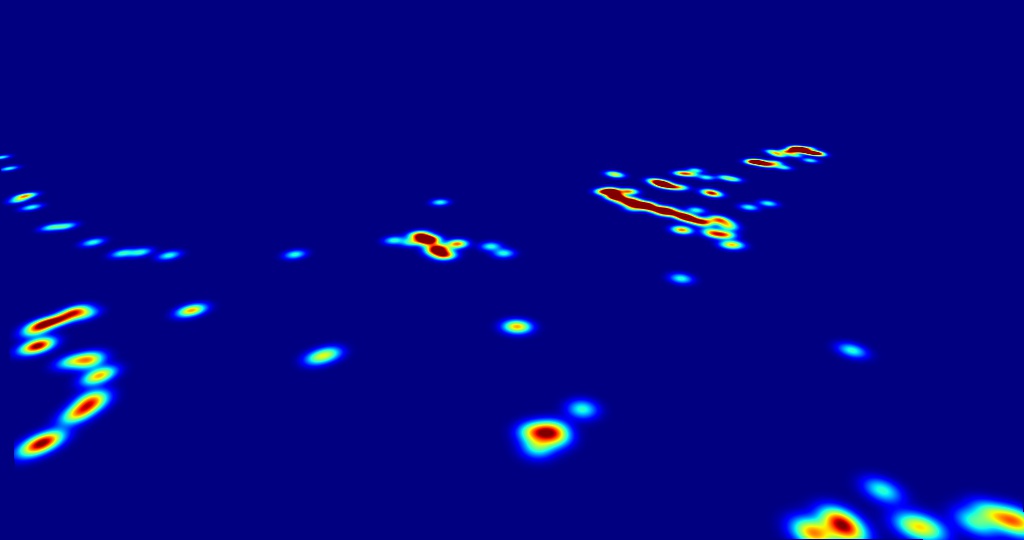}&
    \includegraphics[width=.22\linewidth]{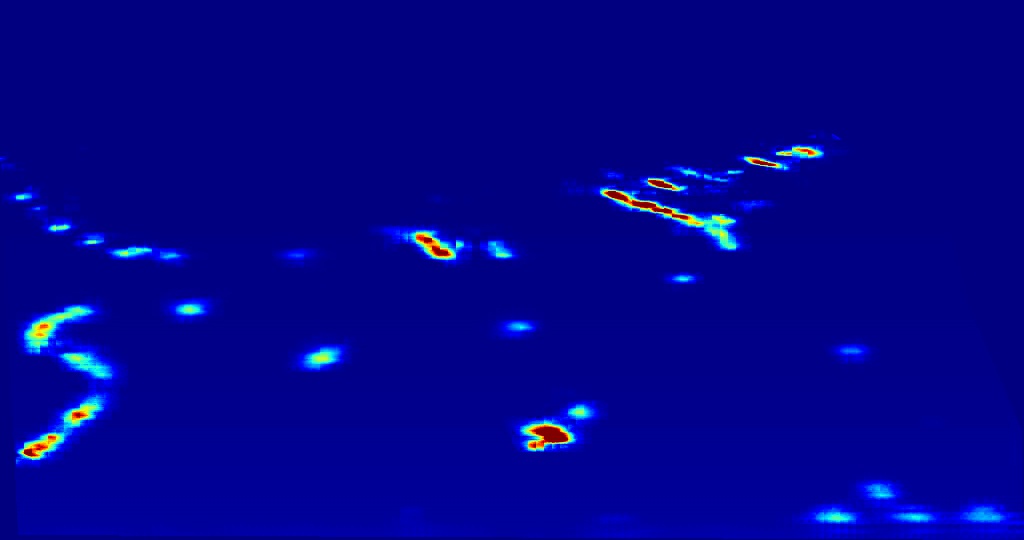}\\
     (a)&(b)&(c)&(d)\\
        \includegraphics[width=.22\linewidth]{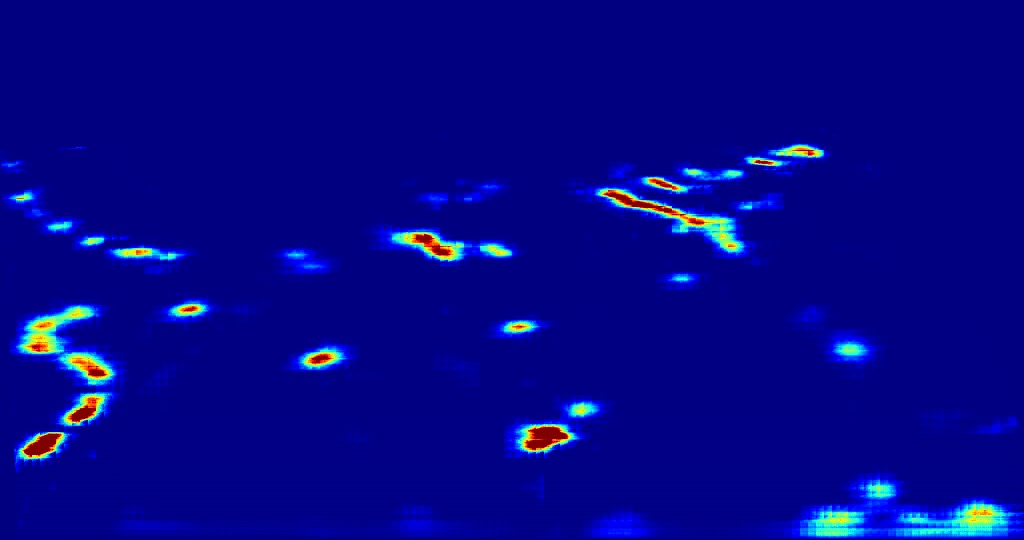}&
        \includegraphics[width=.22\linewidth]{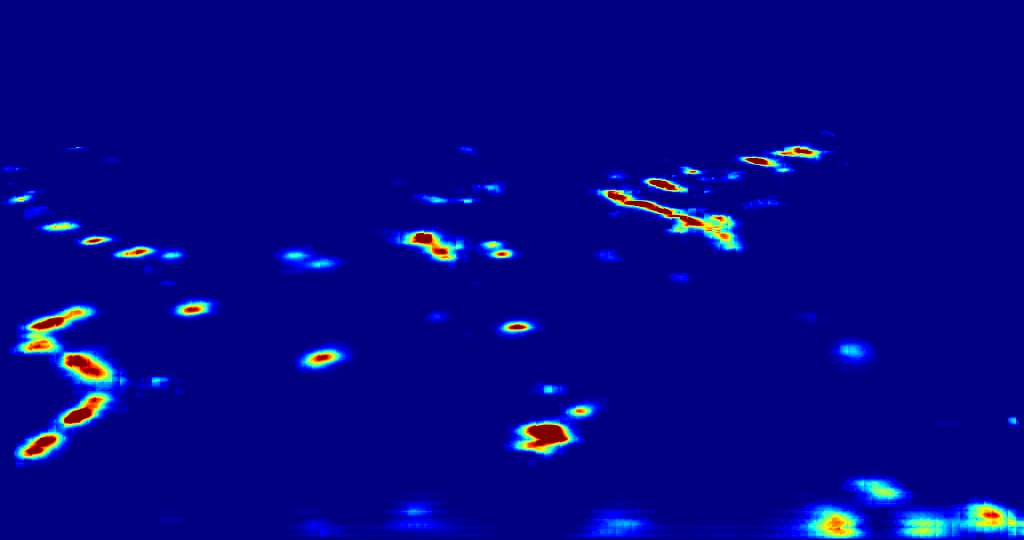}&
        \includegraphics[width=.22\linewidth]{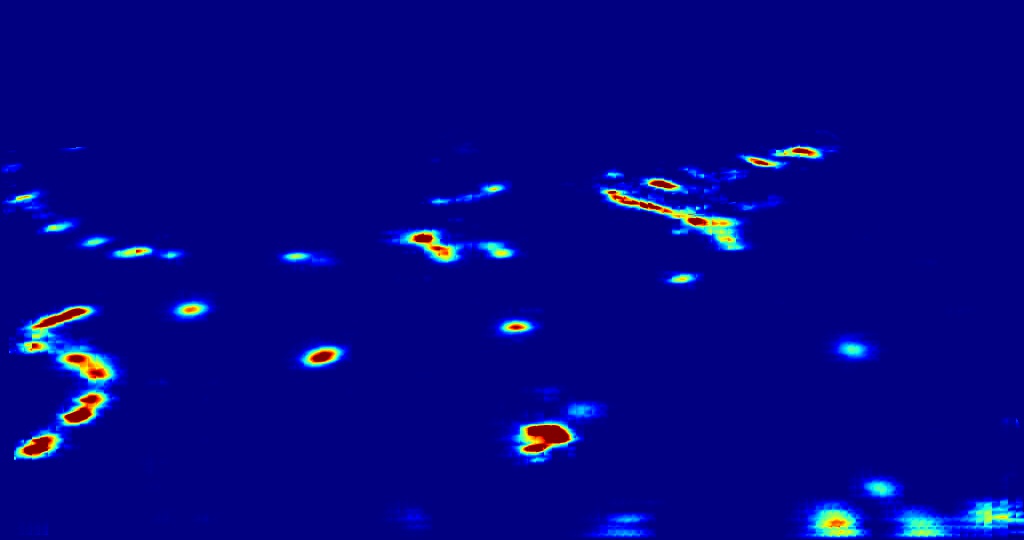}&
          \includegraphics[width=.22\linewidth]{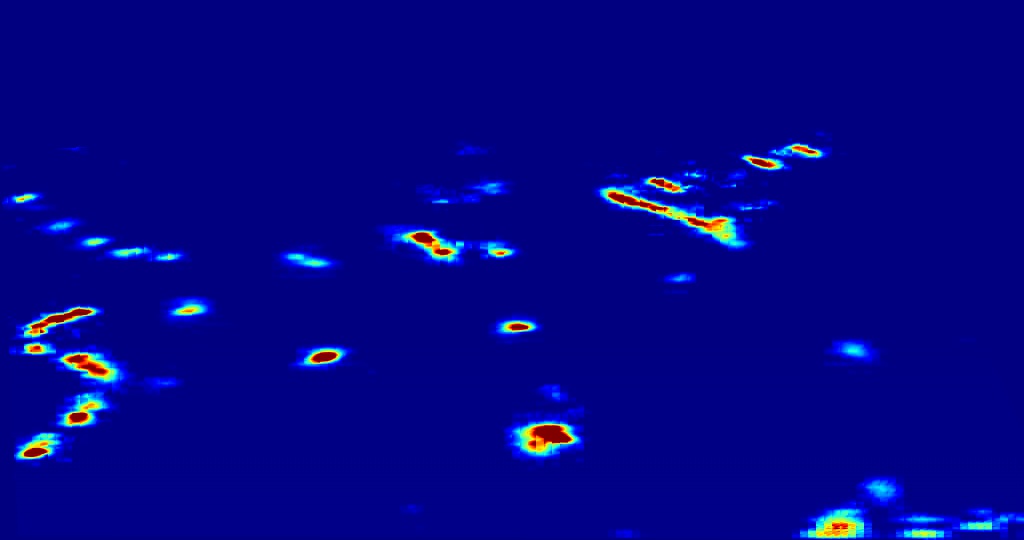}\\
        (e)&(f)&(g)&(h)
  \end{tabular}
  \end{center}
  \vspace{-4mm}
  \caption{{\bf Crowd density estimation on the Campus dataset.} {\small {\bf (a)} Input image. {\bf(b)} $ROI$ overlaid in red. {\bf(c)} Ground truth head plane density.  {\bf(d-h)} Density maps generated by \nogeom{}, \notemp{}, \ours{}(1),   \ours{}(5), and  \ours{}(10).}}
  \label{fig:exp_drone}
  \end{figure*}

We report our comparative results in Tables~\ref{tab:head}, ~\ref{tab:image}, ~\ref{tab:venice_head} and~\ref{tab:venice_image}. Enforcing temporal consistency requires the central frame to be annotated but the other two can be chosen arbitrarily. When running \OURS{}, that is, enforcing both geometry and temporal constraints, we used triplets of images temporally separated by 1, 5, or 10 frames. We provide a qualitative comparison in Fig.~\ref{fig:exp_drone}. 

In Tables~\ref{tab:head} and~\ref{tab:venice_head}, we used Eq.~\ref{eq:1} to convert the image plane densities computed by the baselines into head-plane densities that can be compared to ours.  In Tables~\ref{tab:image} and~\ref{tab:venice_image}, we instead converted our head plane densities into image plane ones that can be compared to theirs. Either way, \notemp{} outperforms the baselines. Furthermore, imposing temporal consistency gives our approach a further boost.


\begin{table}
  \centering
\begin{tabular}{ |p{3cm}|c|c|c|}
  \hline
  Model & $MAE$ & $RMSE$  & $MPAE$ \\
  \hline
  \csr{}~\cite{Li18e} & 50.1 & 54.2 & 125.6 \\
  \mcnn{}~\cite{Zhang16s} & 23.5 & 30.6  & 143.9 \\
  \switch{}~\cite{Sam17} & 91.0 & 120.5 & 330.1 \\
  \hline
  \nogeom{} & 29.2 & 34.8 & 131.2  \\
  \notemp{} & 20.1 & 24.7 & 135.1 \\
  \ours{} (frame interval 1) & \textbf{11.9} & \textbf{15.1} & 116.9 \\
  \ours{} (frame interval 5) & 16.1 & 20.2 & \textbf{113.2}\\
  \ours{} (frame interval 10) & 13.4 & 17.2 & 126.2 \\
  \hline
  \end{tabular}
  \caption{Comparative results in terms of head plane crowd density on the \textbf{Campus} dataset.}
  \label{tab:head}
\end{table}

\begin{table}
  \centering
\begin{tabular}{ |p{3cm}|c|c|c|}
  \hline
  Model & $MAE$ & $RMSE$  & $MPAE$ \\
  \hline
  \csr{}~\cite{Li18e} & 51.3 & 57.6 & 126.4 \\
  \mcnn{}~\cite{Zhang16s} & 24.2 & 37.1  & 146.2 \\
  \switch{}~\cite{Sam17} & 91.7 & 122.1 & 340.7 \\
  \hline
  \nogeom{} & 29.8 & 35.2 & 132.0  \\
  \notemp{} & 21.2 & 24.7 & 136.8 \\
  \ours{} (frame interval 1) & \textbf{12.3} & \textbf{16.0} & 117.3 \\
  \ours{} (frame interval 5) & 16.9 & 22.3 & \textbf{114.1}\\
  \ours{} (frame interval 10) & 14.2 & 18.0 & 128.7 \\
  \hline
  \end{tabular}
   \caption{Comparative results in terms of image plane crowd density on the \textbf{Campus} dataset.}
    \label{tab:image}
\end{table}

\begin{table}
  \centering
\begin{tabular}{ |p{3cm}|c|c|c|}
  \hline
  Model & $MAE$ & $RMSE$  & $MPAE$ \\
  \hline
  \csr{}~\cite{Li18e} & 38.5 & 42.7  &  121.3 \\
  \mcnn{}~\cite{Zhang16s} & 132.7 & 145.3 &  367.6  \\
  \switch{}~\cite{Sam17} &  61.2 & 72.9  & 163.2 \\
  \hline
  \nogeom{} & 36.8  & 39.9 &  115.7 \\
  \notemp{} & 26.1 &35.3 & 107.2\\
  \ours{} (frame interval 1) &  24.8 & 32.7 & 103.2  \\
  \ours{} (frame interval 5) & \textbf{18.2}  & \textbf{26.6} & 98.7 \\
  \ours{} (frame interval 10) & 22.9  & 34.3  & \textbf{94.2} \\
  \hline
  \end{tabular}
   \caption{Comparative results in terms of head-plane crowd density on the \textbf{Venice} dataset.}
  \label{tab:venice_head}
\end{table}

\begin{table}
  \centering
\begin{tabular}{ |p{3cm}|c|c|c|}
  \hline
  Model & $MAE$ & $RMSE$  & $MPAE$ \\
  \hline
  \csr{}~\cite{Li18e} & 39.2 & 44.0  &  124.7 \\
  \mcnn{}~\cite{Zhang16s} & 133.7 & 148.4 &  368.2  \\
  \switch{}~\cite{Sam17} &  63.1 & 75.8  & 165.4 \\
  \hline
  \nogeom{} & 37.2  & 40.4 &  116.3 \\
  \notemp{} & 27.3 &37.2 & 108.9\\
  \ours{} (frame interval 1) &  25.2 & 33.4 & 104.7  \\
  \ours{} (frame interval 5) & \textbf{18.7}  & \textbf{27.0} & 99.2 \\
  \ours{} (frame interval 10) & 23.6  & 35.2  & \textbf{95.1} \\
  \hline
  \end{tabular}
   \caption{Comparative results in terms of image plane crowd density on the \textbf{Venice} dataset.}
  \label{tab:venice_image}
\end{table}

\section{Conclusion}

\begin{figure}
\centering
\begin{tabular}{cccc}
	\includegraphics[width=.21\linewidth]{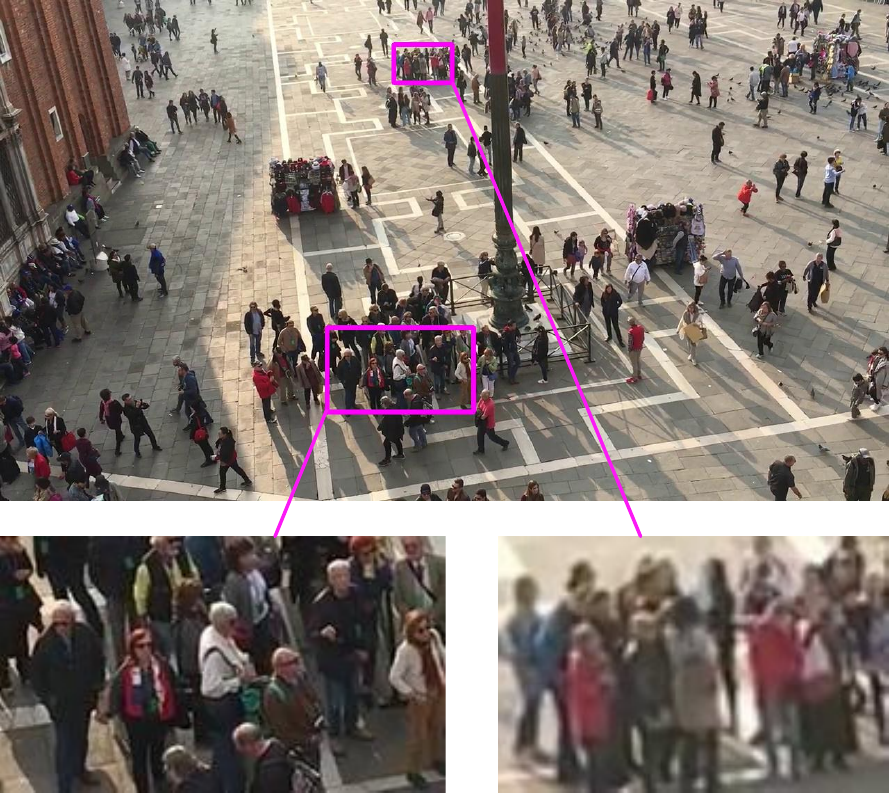}&
	\includegraphics[width=.21\linewidth]{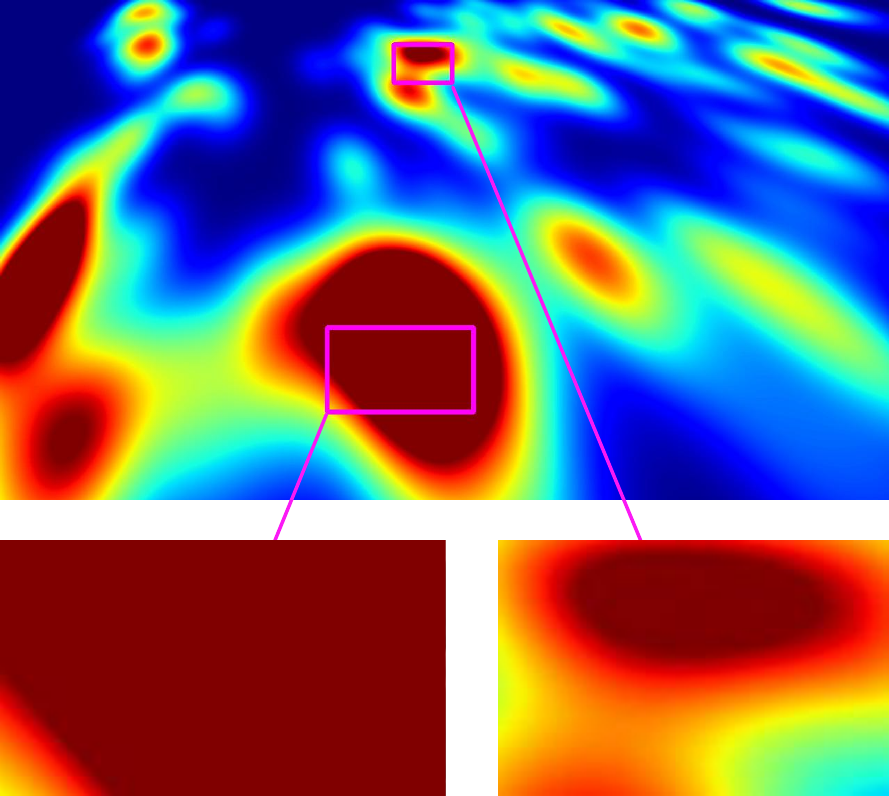}& 
	\includegraphics[width=.21\linewidth]{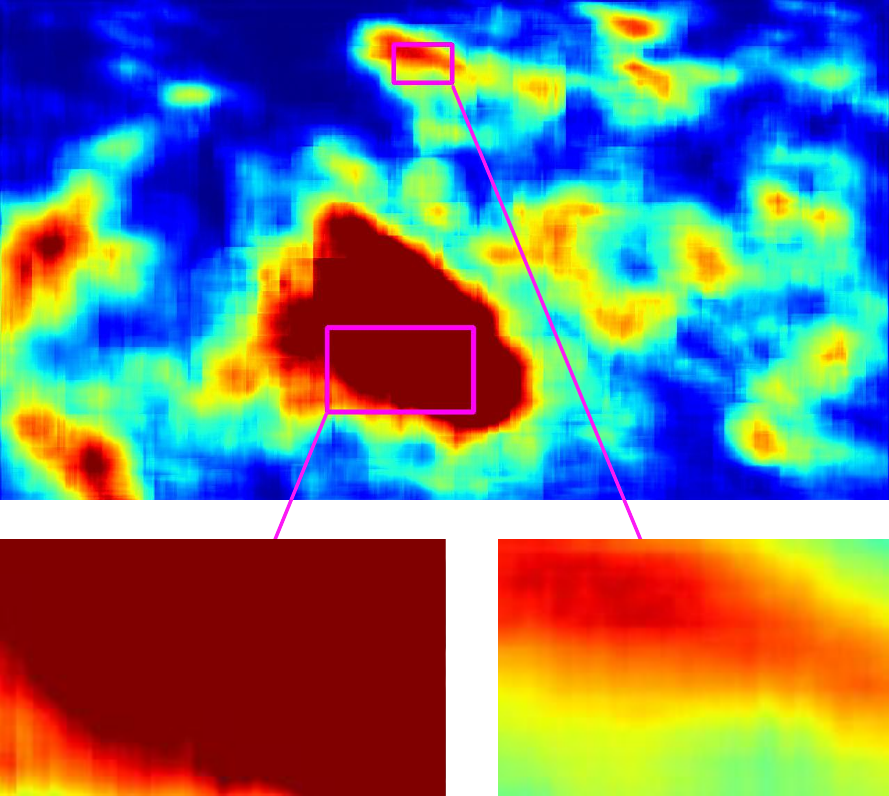} &
	\includegraphics[width=.21\linewidth]{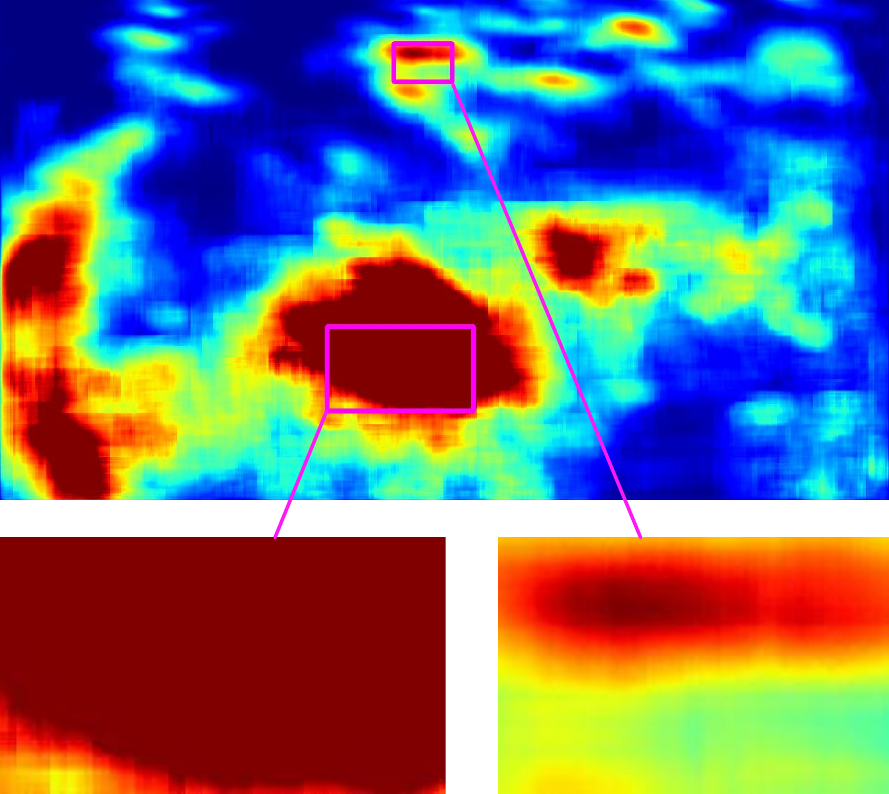}\\
	(a)&(b)&(c)&(d)
\end{tabular}
\vspace{-3mm}
\caption{
	{\bf Crowd density estimation on the Venice dataset.} {\small 
	{\bf (a)} An image of Piazza San Marco in Venice. The two purple boxes highlight patches in which the crowd density per square meter is similar. 
	{\bf (b)} Ground-truth {\it head plane density}.
	{\bf (c)} Density maps generated by \nogeom{}. 
	{\bf (d)} Density maps generated by \ours{}(5). Note that the peak densities are very close, whereas they are quite different in the \nogeom{} version.
}}

\label{fig:loc}
\end{figure}

In this paper, we have shown that providing to a deep net an explicit model of perspective distortion effects as an input, along with enforcing physics-based spatio-temporal constraints, substantially increases performance. In particular, it yields not only a more accurate people count but also a better localization of the high-density areas, as can be seen in Fig.~\ref{fig:loc}.

This is of particular interest for crowd counting from a moving drone that can register its camera with respect to the scene using its internal sensors and therefore estimate the required perspective model. Our approach is equally applicable to mobile device that also possess internal sensors or can use standard photogrammetric techniques to estimate their 3D pose. In future work, we will incorporate this approach into the landing
system of the drone to allow for automated landing in potentially
crowded scenes.


\bibliographystyle{IEEEtran}
\bibliography{short,string,vision,learning}

\end{document}